%% file: main.tex
\documentclass[11pt,a4paper]{article}
\usepackage[hyperref]{emnlp2018}
\usepackage{times}
\usepackage{latexsym}

\usepackage{amssymb}
\usepackage{graphicx}
\usepackage{multirow}
\usepackage{textcomp}
\usepackage{array}
\usepackage{url}
\usepackage{color,comment}

\usepackage[caption=false]{subfig}
\usepackage{enumitem}
\usepackage{booktabs}
\usepackage{amsmath}
\usepackage{cleveref}
\usepackage{rotating}
\usepackage{stackengine}

\usepackage{url}

\aclfinalcopy 

\usepackage{todonotes}
\newcounter{hfang}

\newcounter{chenghao}
\title{Robust cross-domain disfluency detection with pattern match networks}

\author{Vicky Zayats \\
  Electrical Engineering Department \\
  University of Washington \\
  {\tt vzayats@uw.edu} \\\And
  Mari Ostendorf \\
  Electrical Engineering Department \\
  University of Washington \\
  {\tt ostendor@uw.edu} \\}

\date{}

\begin{document}
\maketitle

\begin{abstract}
In this paper we introduce a novel pattern match neural  network  architecture  that  uses  neighbor similarity scores as features,
eliminating the need for feature engineering in a disfluency detection task.  We evaluate the approach in disfluency detection for four different speech genres,
showing that the approach is  as  effective  as   hand-engineered pattern match features when used on in-domain data and achieves superior  performance in  cross-domain  scenarios. 
\end{abstract}

\section{Introduction}

\input{intro.tex}

\section{Method}\label{sec:approach}

\input{method.tex}

\section{Experiments and Analysis}

\input{experiments.tex}

\section{Related Work}

\input{related.tex}

\section{Conclusion}\label{sec:conclusion}

\input{conclusions.tex}

\newpage
\bibliographystyle{acl_natbib}
\bibliography{disfluency}

\end{document}

%% file: intro.tex
Disfluencies are self corrections in spontaneous speech, including filled pauses, repetitions, repairs and false starts.
Below are some examples of disfluent sentences from the corpora used in this work:
\begin{quote}
\begin{center}
\vspace{-0.2cm}
\hspace*{-0.4cm} \textit{[He is + our clients are] subject to }

\vspace{-0.15cm}
\hspace*{-0.4cm} \textit{[it 's + \{you know\} it 's] one of the last}



\vspace{-0.15cm}
\hspace*{-0.4cm} \textit{you want [it + just something] that is}
\vspace{-0.1cm}
\end{center}
\end{quote}
The brackets indicate the beginning and of the disfluency and the end of the correction. The reparandum includes the words before the interruption point (+), which the speaker intends to replace or ignore. 
The words that come as a correction to the reparandum follow the interruption point. An optional interregnum (in brackets) follows the interruption point, including words such as filled pauses, discourse markers, etc. 
Systems are usually evaluated on the ability to correctly identify the reparandum. 

Previous studies on disfluency detection observe that a repair is often a ``rough copy" of a reparandum \cite{charniak01,zwarts+10}; thus, hand-crafted pattern match features play an important role in many disfluency detection approaches. They have been shown to be helpful to both sequential and parsing based approaches \cite{wang2017transition,zayats2016,semi-markov,wu2015efficient,qian+13}. In the examples above, ``he" resembles ``clients", ``is" resembles ``are" and ``it 's" is a repetition. However, in many cases, the pattern match is not simple, if present at all, as in the last example.
In addition, disfluencies can have domain-dependent characteristics. 

In this work. we present a novel architecture that allows automated discovery of the patterns instead. We first calculate a similarity score between neighboring words in a sentence. Then, we use those scores directly to identify multi-token patterns with convolutional neural networks (CNN). Experiments show that our proposed architecture has an in-domain performance comparable to using hand-crafted pattern match features, and it outperforms baselines in cross-domain setting. In this paper, our main contribution is a novel neural network architecture which allows automatic discovery of patterns using a mechanism similar to attention but where the similarity scores serve as
input features to a CNN, rather than as weights for computing a context vector. 
In addition to eliminating the need for feature engineering, the model is shown to be robust in cross-domain testing.

%% file: method.tex
\begin{figure*}[t]
\includegraphics[scale=0.37]{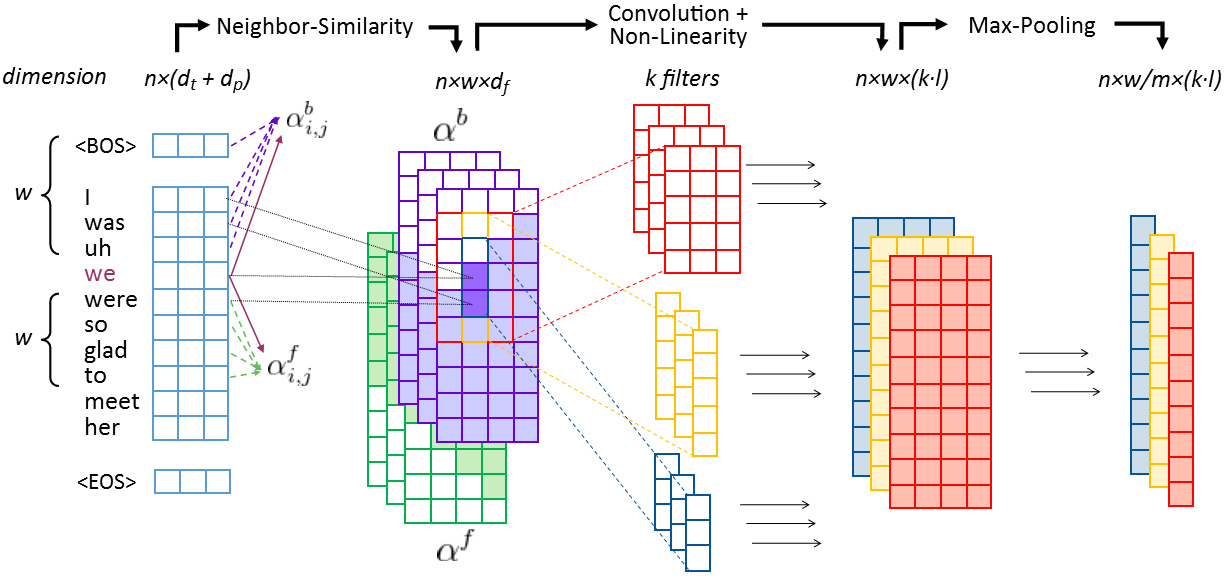}
\centering
\vspace{-0.8cm}
\caption{An illustration of the model. In this example, the backward neighbor-similarity layer $\alpha^b$ identifies that \textit{``we"} has high similarity with \textit{``I"} and \textit{``were"} has high similarity with \textit{``was"}. Both \textit{``we"} and \textit{``were"} are at a distance of 3 from the corresponding \textit{``I"} and \textit{``was"}. A convolutional filter can catch the horizontal pattern in row 3, thus indicating the presence of a bigram pattern match between \textit{``we were"} and \textit{``I was"}. }
\vspace{-0.4cm}
\label{fig:model}
\end{figure*}

The main motivation behind our approach is to allow the model to automatically learn and find patterns in sentences without defining them via hand-crafted features. Our proposed model uses two levels to automatically find patterns in sentences. In the first level we calculate similarities for each word in a sentence with words in the surrounding window, which we refer to as neighbor similarity. After calculating the single-token similarity weights, in the second level, we use those weights as features to extract local patterns using a convolutional neural network. The schematic diagram of the model is presented in Figure \ref{fig:model}.


\subsection{Neighbor Similarity}

The hand-crafted pattern match features used in disfluency detection are usually in the form of ``\textit{does the exact word/POS/bigram appeared previously in a fixed length window?}" In our work, instead of  manually defining similarity functions (e.g.\ exact match of the word/POS), we learn similarity functions between individual words in a sentence. For each word in a sentence of length $n$, we calculate a similarity between the given word $x_i$ and each of the words $x_j$ in the preceding/following window of size $w$, for $j \in [i\pm 1, \dots, i\pm w]$. In our task we used cosine similarity $sim$ to calculate the alignment score between a pair of words due to the straightforward resemblance of words in the reparandum and repair:
\vspace{-0.2cm}
\begin{equation}
 \alpha^{\{f,b\}}_{i,j} = sim(W^{1}x_i, W^{2}x_j)   
\end{equation}
%
where $W^1,W^2 \in \mathbb{R}^{d_f\times d_g \times(d_t+d_p)}$ are learned.

We refer to similarity scores in the preceding/following windows as $\alpha^b$ (backward) and $\alpha^f$ (forward), respectively. For the cases when the $x_j$ is outside of the sentence boundaries, we set the similarity score to be zero. In order to capture multiple types of similarities between two word representations, we concatenate token and part-of-speech (POS) embeddings and learn multi-dimensional similarity scores $\alpha_{i,j} \in \mathbb{R}^{d_f}$. 
In our experiments, we set $d_g=(d_t+d_p)$, where $d_t$ and $d_p$ are the dimensions of token and POS embeddings, respectively. 
The overall dimension of the similarity matrix is $\alpha \in \mathbb{R}^{w\times n \times d_f}$, where $n$ is the sentence length and $w$ is the size of the window. 

\begin{table*} [t]
\begin{center}
\begin{tabular}{|l|c|c|c|c|c|c|c|}
\hline \bf Model  & pattern& SWBD test  & CallHome & SCOTUS & FCIC \\ \hline \hline
CRF & \textendash & 71.3 & 58.1 & 70.8 & 53.2 \\
    &  \checkmark  & 82.5 & 63.2 & 79.2 & 63.8 \\ \hline
LSTM  & \textendash & 82.9 & 54.1 & 57.9 & 36.9 \\
    & \checkmark  & \bf 86.8 & 58.7 & 66.6 & 48.9 \\ \hline
LSTM + sim  & \textendash & 85.9 & 64.8 & 78.8 & 65.0 \\ \hline
LSTM + sim + conv  & \textendash & 86.7 & \bf 65.2 & \bf 79.9 & \bf 66.1 \\ \hline
\end{tabular}
\end{center}
\vspace{-0.3cm}
\caption{\label{tab:results}F1 scores on cross-domain disfluency detection. ``pattern" stands for hand-crafted pattern match features.}
\vspace{-0.4cm}
\end{table*}

\begin{table} [t]
\begin{center}
\begin{tabular}{|l|c|c|c|}
\hline \bf Model & CallHome &  SCOTUS & FCIC\\ \hline
CRF  & 71.5 & 90.9 & 88.7 \\ \hline
LSTM  & 54.7 & 71.8 & 58.7 \\ \hline
LSTM pm net  & 68.4 & 88.0 & 87.2 \\ \hline
\end{tabular}
\end{center}
\vspace{-0.3cm}
\caption{\label{tab:prec_rec} Precision across domains for CRF, LSTM with hand-engineered pattern match features and LSTM with pattern match networks.}
\vspace{-0.4cm}
\end{table}

\begin{table*}
\begin{center}
{\small
\begin{tabular}{|l|c|p{5.3in}|}\hline
\bf Corpus & \bf Ex & \bf Sentence\\ \hline
CallHome & 1 & Oh he [looks like + John Travolta but he has like] curly blond hair. \\
         & 2 &[I do n't think + [I know her + but I 've]] heard of her \\ \hline
SCOTUS   & 3 & What is your [authority + for that proposition, Mr. Guttentag, your case authority]? \\
         & 4 & \dots as to permit review in [the court + of appeals, then the district court] habeas corpus procedure need \dots \\ \hline
FCIC     & 5 &Thank you for the opportunity [to + contribute to the commission 's work to] understand the causes of \dots\\ 
         & 6 & \dots counter parties were unaware [of + the full extent of] those vehicles and therefore could not \dots \\ \hline
\end{tabular}
}
\end{center}
\vspace{-0.3cm}
\caption{Example sentences wrongly predicted as disfluent by LSTM model with hand-crafted pattern match features. The brackets indicate predicted disfluency regions, where the respected gold annotation is ``non-disfluent".  \vspace{-0.6cm}
\label{tab:examples}}
\end{table*}

\subsection{Convolution over Similarity Features}
While neighbor-similarity features can be useful, they do not exploit all the information about repeating patterns. A simple example can be a bigram pattern match feature:
the model can find a similarity between closely related words on the unigram level, but it is unable to directly identify cases where the bigram would be repeated. 
To capture temporal patterns presented in neighbor-similarity scores,  we apply convolutional filters on the output of the neighbor-similarity layer, followed by a non-linearity ($\tanh$). For example, in Figure \ref{fig:model}, the neighbor-similarity layer would identify similarity between individual tokens \textit{``we"} and \textit{``I"}, and \textit{``were"} and \textit{``was"}. A convolutional layer on top would capture the horizontal bigram pattern between \textit{``we were"} and \textit{``I was"}. 
The output of the convolutional layer is $f_{conv}(\alpha) \in \mathbb{R}^{w\times n \times k l}$, where $k$ is number of different filter shapes and $l$ is the number of output filters for each filter shape. We apply the max-pooling layer with a downsample rate $m$ on top to summarize the convolutional layer output at each time $i$. 
The output of the max-pooling layer is $g(\alpha) \in \mathbb{R}^{w / m\times n \times k l}$. We flatten the outputs of the max-pooling layer and concatenate with the input feature embeddings,
and input the resulting vector to an LSTM.

%% file: experiments.tex
Our experiments target both in-domain and cross-domain scenarios. In addition, we analyze the differences in errors made by the models.

\subsection{Data}
Switchboard (SWBD) \cite{Switchboard} is the standard and largest corpus used for disfluency detection. 
The current state-of-the-art in disfluency detection achieves F1 score of 88.1 on the SWBD test set \cite{wang2017transition}. 
In addition to Switchboard, we test our models on three out-of-domain publicly available datasets  annotated with disfluencies \cite{zayats2014multidomain}: 

\hspace{-0.38cm}\textbf{Switchboard}: phone conversations between strangers on predefined topics;

\hspace{-0.38cm}\textbf{CallHome}: phone conversations between family members and close friends; 

\hspace{-0.38cm}\textbf{SCOTUS}: transcribed Supreme Court oral arguments between justices and advocates;

\hspace{-0.38cm}\textbf{FCIC}: two transcribed hearings from Financial
Crisis Inquiry Commission.

\subsection{Model Comparisons}
We train the CRF \footnote{\url{https://taku910.github.io/crfpp}} and 
bidirectional LSTM-CRF\footnote{\url{https://github.com/UKPLab/emnlp2017-bilstm-cnn-crf}} models as baselines, both with and without pattern match features. For simplicity, we refer to bidirectional LSTM-CRF model as just LSTM.
In all models we use \textbf{identity features}, including word, POS tag, whether the word is a filled pause, discourse marker, edit word or fragment. The \textbf{hand-crafted pattern match features} include: distance to the repeated \{word, bigram, POS, word+next POS, POS bigram, POS trigram\} in the \{preceding, following\} window; whether word bigram is repeated in the \{preceding, following\} window allowing some words in between the two words; and distance to the next conjunction word. Following \cite{zayats2016}, we use 8 BIO states. 
For each experiment, we average the performance of 15 randomly initialized models.

For our proposed model we use the following parameters: window size $w=10$, neighbor-similarity dimension $d_f=100$, $k=5$ different filter shapes: $[1,1],[3,1],[3,3],[5,1]$ and $[5,3]$, with output filter dimension $l=16$, and downsample rate $m=3$. In our initial experiments we have tuned the parameters $d_f,k,l$ and $m$ to the values mentioned above using the Switchboard development set.

\subsection{Results}

The cross-domain experiment results are presented in Table \ref{tab:results}. In general, for the in-domain data (Switchboard), the pattern match networks achieve performance comparable to the LSTM model with hand-crafted pattern match features, and significantly outperforms the CRF model. In addition, our model is robust compared to the baselines when applied to out-of-the-domain data, with a consistent improvement over the CRF. Surprisingly, the LSTM performs poorly on out-of-the-domain data. 
To better understand the model differences, in the next section we conduct error analysis and discuss the findings.

\subsection{Error analysis}

The difficulty in applying the model on out-of-domain data lie in both difference in corpora and underlying model. There is substantial variation in vocabulary, conversational style, disfluency types, and sentence segmentation criteria across corpora. CallHome is more casual than SWBD; SCOTUS and FCIC are formal high stakes discussions with vocabularies highly dissimilar to SWBD. SCOTUS, FCIC, and CallHome contain 2, 5 and 7 times more restarts token-wise, respectively, than Switchboard. Also, on average, disfluencies in all three out-of-the-domain corpora tend to be longer, especially in CallHome and FCIC. 

To further study the effect of pattern match features, we trained models with identity features only. 
When comparing models with identity features only, CRF performs poorly compared to LSTM on in-domain data. On the other hand, LSTM with no pattern match features performs considerably well on in-domain Switchboard. By looking at cross-domain results, we see that CRF is more stable across the domains, compared to the LSTM. We hypothesize that LSTM is more powerful in learning specific data structure, compared to the CRF, and overfit the models to match Switchboard style. On the other hand, pattern match networks are better at capturing patterns that are more general across domains. 

Table \ref{tab:prec_rec} shows that that there is a significant drop in precision for LSTMs with hand-derived features. In particular, the LSTM with pattern match features ``hallucinates" a lot of disfluencies, longer ones in particular. This might be due to the long memory of the LSTM, as opposed to CRF, which tends to be more local. 
Table \ref{tab:examples} presents some examples with false positives made by the LSTM with hand-engineered features, but were correctly identify by our model. 


%% file: related.tex
Most work on disfluency detection fall into three main categories: sequence tagging, noisy-channel and parsing-based approaches.
Sequence tagging approaches include conditional random fields (CRF) \cite{georgila09,ostendorf+13,zayats2014multidomain}, Max-Margin Markov Networks (M$^3$N) \cite{qian+13}, Semi-Markov CRF \cite{semi-markov}, and 
recurrent neural networks \cite{disfluency_rnn,zayats2016,wang2016neural}. The main benefit of sequential models is the ability to capture long-term relationships between reparandum and repairs.
Noisy channel models operate on a relationship between the reparandum and
repair for identifying disfluencies \cite{charniak01,zwarts+10}. \citet{lou2017disfluency} used a neural language model to rerank
sentences using the noisy channel model. 
Approaches that combine parsing and disfluency removal tasks include \cite{rasooli2013joint,honnibaljoint,tran2017joint}. The current state-of-the-art in disfluency detection uses a transition-based neural model architecture \cite{wang2017transition}.

There exist a limited effort on cross-domain disfluency detection. \citet{georgila10} used CRF and integer linear programming in detecting disfluencies in human-agent interactions.
\citet{zayats2014multidomain} introduced pattern match features with a CRF and released the datasets that we use for testing in our experiments.  \citet{zayatsunediting} used semi-supervised learning in adapting the model to SCOTUS non-careful transcripts.

%% file: conclusions.tex
In this paper we introduce a novel neural network architecture which allows automatic discovery of patterns and directly uses similarity scores as
input features to a CNN. We show that our approach can be as effective as using carefully designed, hand-engineered pattern match features in a disfluency detection task, eliminating the need for feature engineering, and show that it is robust in cross-domain testing.
In the future, following \citet{ganin2014unsupervised}, we are interested in exploring domain adaptation techniques in disfluency detection. Motivated by \citet{tran2017joint}, we are also interested in incorporating prosody information to further improve disfluency detection.